\documentclass[sigconf, nonacm]{acmart}

\usepackage[ruled,vlined]{algorithm2e}
\usepackage{tabularx}

\usepackage{soul}
\usepackage{url}
\usepackage{hyperref}
\usepackage[utf8]{inputenc}
\usepackage{caption}
\usepackage{graphicx}
\usepackage{booktabs}
\usepackage{amsmath, bm}
\usepackage{bbm}
\usepackage{enumitem}
\usepackage{natbib}
\usepackage{amsthm}
\usepackage{subcaption}
\usepackage{comment}
\usepackage{xcolor}
\usepackage{siunitx}
\usepackage{mathtools}
\usepackage{multirow}
\DeclareMathOperator*{\sometext}{\Big\Vert}

\AtBeginDocument{%
  \providecommand\BibTeX{{%
    \normalfont B\kern-0.5em{\scshape i\kern-0.25em b}\kern-0.8em\TeX}}}

\setcopyright{acmcopyright}\acmConference[]{arXiv Preprint}
\copyrightyear{2023}
\acmYear{2023}
\acmDOI{XXXXXXX.XXXXXXX}

\acmPrice{15.00}
\acmISBN{978-1-4503-XXXX-X/18/06}

\begin{document}



\title{Towards Assumption-free Bias Mitigation}

\author{Chia-Yuan Chang}
\affiliation{%
  \institution{Texas A\&M University}}
\email{cychang@tamu.edu}

\author{Yu-Neng Chuang}
\affiliation{%
  \institution{Rice University}}
\email{ynchuang@rice.edu}

\author{Kwei-Herng Lai}
\affiliation{%
  \institution{Rice University}}
\email{khlai@rice.edu}

\author{Xiaotian Han}
\affiliation{%
  \institution{Texas A\&M University}}
\email{han@tamu.edu}

\author{Xia Hu}
\affiliation{%
  \institution{Rice University}}
\email{xia.hu@rice.edu}

\author{Na Zou}
\affiliation{%
  \institution{Texas A\&M University}}
\email{nzou1@tamu.edu}




\begin{abstract}

Despite the impressive prediction ability, machine learning models show discrimination towards certain demographics and suffer from unfair prediction behaviors. To alleviate the discrimination, extensive studies focus on eliminating the unequal distribution of sensitive attributes via multiple approaches. However, due to privacy concerns, sensitive attributes are often either unavailable or missing in real-world scenarios. Therefore, several existing works alleviate the bias without sensitive attributes. Those studies face challenges, either in inaccurate predictions of sensitive attributes or the need to mitigate unequal distribution of manually defined non-sensitive attributes related to bias. The latter requires strong assumptions about the correlation between sensitive and non-sensitive attributes. As data distribution and task goals vary, the strong assumption on non-sensitive attributes may not be valid and require domain expertise.
In this work, we propose an assumption-free framework to detect the related attributes automatically by modeling feature interaction for bias mitigation. The proposed framework aims to mitigate the unfair impact of identified biased feature interactions.
Experimental results on four real-world datasets demonstrate that our proposed framework can significantly alleviate unfair prediction behaviors by considering biased feature interactions.
Our source code is available at: \href{https://anonymous.4open.science/r/fairint-5567}{\texttt{https://anonymous.4open.science/r/fairint-5567}}
\end{abstract}

\maketitle
\section{Introduction}\label{sec:intro}

Machine learning models have shown superiority in various high-stake decision-makings~\cite{devlin2018bert, krizhevsky2012imagenet, wang2020skewness, vashishth2019composition,mehrabi2021survey, du2020fairness, chuang2020tpr, wan2021modeling}, and have been deployed in many real-world applications, such as credit scoring~\cite{bono2021algorithmic}, loan approval~\cite{mukerjee2002multi}, criminal justice~\cite{heidensohn1986models,green2018fair}, education opportunity~\cite{friedler2016possibility}.
However, machine learning models show discrimination towards certain demographics and suffer from biased prediction behavior, which may negatively impact the minority groups in those application fields.
For example, COMPAS, a recidivism prediction system, shows discrimination towards African-American offenders with a higher possibility of becoming a recidivist two years after leaving prison~\cite{angwin2016machine}. Recent works focus on bias mitigation techniques to alleviate discrimination in machine learning models.


Existing works to tackle the fairness issues are generally based on two groups of assumptions, i.e., \textit{bias assumptions} and \textit{correlation assumptions}. 
For the works based on \textit{bias assumptions}, they mitigate bias with known distributions of sensitive attributes by fairness regularization~\cite{kamishima2011fairness, kamishima2012fairness, berk2017convex, beutel2019putting, beutel2019fairness, agarwal2019fair, aghaei2019learning, choi2019can, di2020counterfactual, jiang2020wasserstein, nam2020learning, chuang2022mitigating}, contrastive learning~\cite{dwork2012fairness, shen2016disciplined, zafar2017fairness, kusner2017counterfactual, kearns2018preventing, hashimoto2018fairness, zafar2019fairness, bose2019compositional, garg2019counterfactual, kobren2019paper, cheng2021fairfil, zhou2021contrastive}, adversarial learning~\cite{edwards2015censoring, zhang2018mitigating, xu2018fairgan, madras2018learning, wadsworth2018achieving, elazar2018adversarial, sweeney2020reducing}, disentanglement representations~\cite{locatello2019fairness, shen2020interfacegan, kim2021counterfactual, park2021learning}, and representation neutralization~\cite{du2021fairness}.
However, due to privacy concerns, sensitive attributes are often missing~\cite{hashimoto2018fairness, coston2019fair, lahoti2020fairness, yan2020fair, liu2021just}. Therefore, existing works adopt clustering methods~\cite{yan2020fair} and an auxiliary module~\cite{lahoti2020fairness} to simulate the sensitive attributes. However, they often suffer from the inaccuracy of the predicted sensitive attributes when adopting clustering algorithms~\cite{lahoti2020fairness}.
Thus, the work based on \textit{correlation assumptions}, FairRF~\cite{zhao2022towards}, addresses the unfair issues with a strong assumption that the unfair model prediction actually comes from the relationship between sensitive attributes and a set of predefined related non-sensitive attributes.

\begin{figure}[t!]
\centerline{\includegraphics[width=0.48\textwidth]{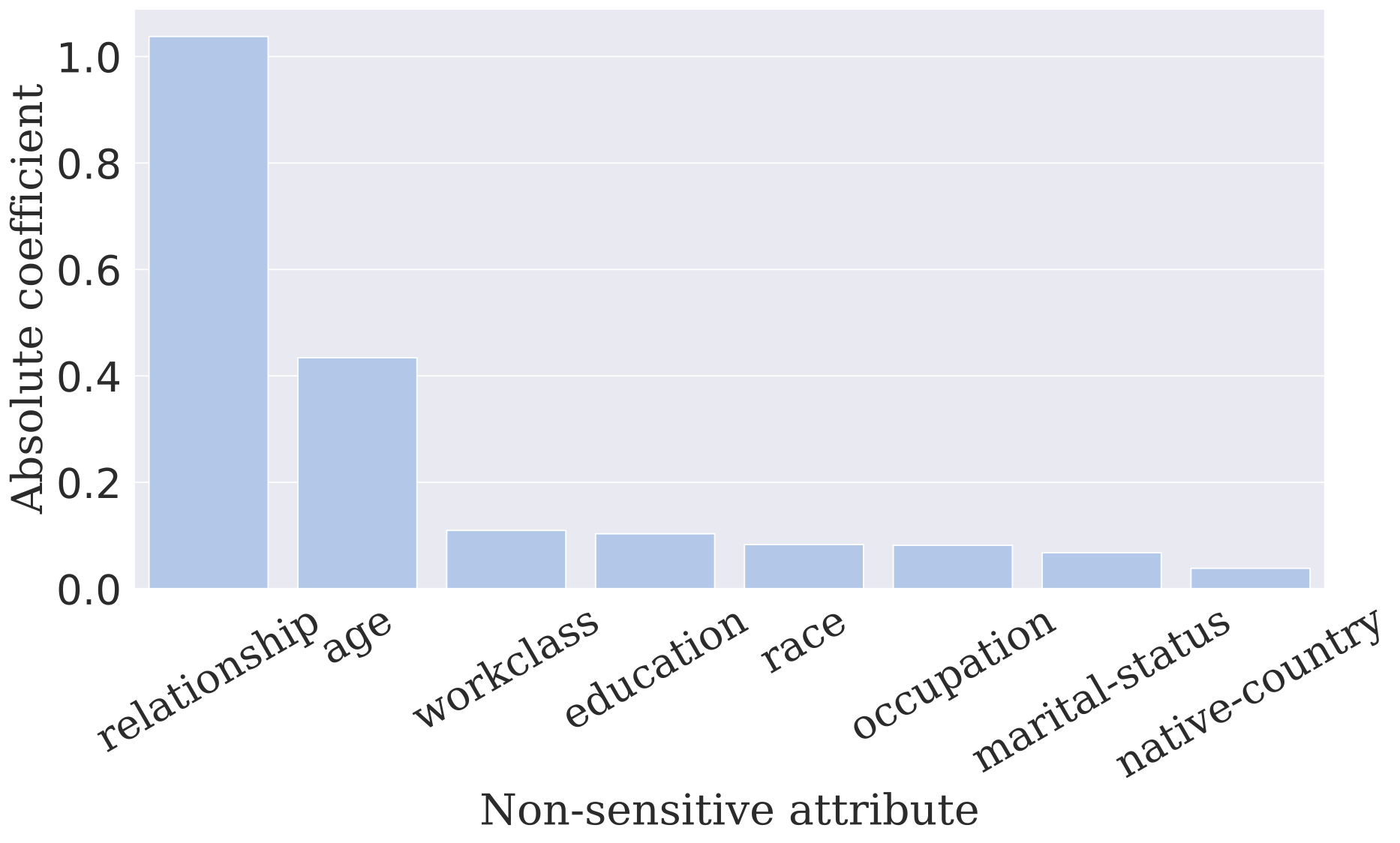}}
\caption{Coefficient of a linear model predicting sensitive attribute "gender" on the Adult dataset, we consider which as the relationship between "gender" and all other features.}
\label{fig:linear_params}
\end{figure}

In this paper, we argue that \textit{correlation assumptions} between sensitive and non-sensitive attributes may not be valid as data distribution and task goals vary. For example, FairRF~\cite{zhao2022towards} predefines (inherently assumes) that the related features of \textit{gender} in the Adult dataset are \textit{age}, \textit{relationship}, and \textit{marital status}. To show this assumption is invalid, we conducted an experiment to explore the relationship between \textit{gender} and all other features. This assumption is not consistent with the linear relationships between \textit{gender} and all other features, as shown in Figure~\ref{fig:linear_params}. Additionally, domain expertise and knowledge are required to predefine the related features.
Therefore, we raise the following question: \textbf{Can we achieve fairness without assuming a predefined relation between sensitive and non-sensitive attributes?}

To tackle the limitations of \textbf{1)} the correlation assumption that unfair model prediction comes from the handcrafting predefined related attributes and \textbf{2)} the further fairness problems caused by feature interactions, we aim to develop an assumption-free framework to automatically detect and integrate feature interactions for bias mitigation.
It is nontrivial to achieve our goal due to the following challenges.
First, in the real-world scenario, implicit bias of feature interactions are difficult to be detected, especially when sensitive attributes are missing. 
Specifically, it is hard to find the high-order statistical interactions that may lead to biased predictions of deep neural networks due to the complex model structures.
Thus, when neither the sensitive attributes are available nor make strong correlation assumptions on related features, it becomes very challenging to identify the biased feature interactions.
For example, identifying the biased feature interactions among all the combinations of features without the sensitive attributes may lead to numerous candidate features interactions, which make models extremely hard to learn the distribution of actual biased feature interactions.
Second, it is challenging to mitigate bias in feature interactions due to the uneven distribution among feature interactions. 
For example, without considering the potential uneven distribution of the feature interactions, trained prediction models may fail to detect and mitigate the bias in feature interactions.

To address the aforementioned challenges, we propose FairInt, an assumption-free framework to automatically identify and further mitigate the bias in feature interactions.
Specifically, we develop a sensitive attribute reconstructor for tackling a situation where sensitive attributes are unavailable during the inference stage.
By designing a sensitive-oriented attention score, we develop a biased interaction detection layer to automatically identify the biased feature interactions and then embed the biased interaction information into the latent representation.
It is different from traditional deep neural networks that model feature interactions among all possible feature combinations and cannot identify specific biased feature interactions.
To equalize the probability distribution of sensitive attributes, we design two bias regularizations for debiasing the latent representation that contains biased interaction information.
These two regularizations debias the feature interactions by minimizing the divergence of latent space and the model predictions between different sensitive attribute groups.
We evaluate our framework on four real-world datasets across three different application domains, which include finance, education, and healthcare.
Compared with baseline models, the experimental results demonstrate that the FairInt can successfully further mitigate the biased prediction behaviors while providing similar performances of downstream tasks by considering biased feature interactions.
Moreover, by observing the modeled feature interaction, the FairInt shows the ability to provide better explainability via the designed sensitive-oriented attention score. We highlight our contributions as follows:

\begin{itemize}[leftmargin=*]
     \item We argue that the related attributes with high correlations to sensitive attributes that can be identified by prior knowledge is problematic. Because the correlations between sensitive and non-sensitive attributes will be changed with different models.
	 \item We propose an assumption-free framework to automatically identify and further mitigate the biased feature interactions. Our framework does not need to handcraft related attributes for mitigating the unfair model prediction that comes from the interactions between sensitive and non-sensitive attributes. Instead, the proposed framework automatically identifies related attributes without prior knowledge during the inference stage.
	 \item Experimental results on several real-world datasets demonstrate the effectiveness of the proposed FairInt framework.
  Additionally, our framework provides better explainability via observing the attention weights between sensitive and non-sensitive attributes.
\end{itemize}

\section{Preliminaries}
In this section, we introduce the existing bias mitigation strategies for deep neural networks and feature interaction modeling methods that inspire our proposed framework.

\vspace{-0.1cm}
\subsection{Bias Mitigation}\label{sec:bias_mit}


To tackle the prejudicial decisions problem in deep learning models, there is increased attention to bias mitigation methods in recent studies~\cite{kamishima2012fairness, beutel2019putting, zhang2018mitigating, du2021fairness}.
Extensive approaches apply regularization-based methods to the objective function of the proposed models, which require pre-hoc assumptions to develop.
Existing alleviating techniques are generally based on two groups of assumptions. 

\textbf{Bias Assumptions.}
Because machine learning models show discrimination towards certain demographics, people assume that machine learning models have biased behaviors against certain groups.
With a known distribution of a sensitive attribute set, there are several advancements proposed to mitigate bias, such as:
\textbf{1)} Fairness regularization:  the objective function of the bias mitigated models generally adds the fairness-related constraint terms~\cite{kamishima2012fairness}, which may penalize the prejudiced behaviors of the prediction models. Another existing work~\cite{beutel2019putting} compares the distributions of model predictions of different sensitive attributes and then minimizes KL-divergence between each sensitive attribute.
\textbf{2)} Adversarial learning: adversarial learning alleviates the biased effects from the known sensitive attributions by simultaneously building an \textit{Adversary} with the \textit{Predictor} of machine learning models.
One previous work~\cite{zhang2018mitigating} aims to leverage bias alleviation by proposing an adversarial learning strategy with the given distribution of sensitive attributes.
The model includes a \textit{Predictor}, which accomplishes the downstream task predictions, and an \textit{Adversary}, which predicts the target sensitive attributes. 
The framework adopts adversarial training by minimizing \textit{Predictor} and maximizing adversary, which aims to debias the unfair situations brought from \textit{Predictor}.
\textbf{3)} Latent representation neutralization: one latent representation neutralization work~\cite{du2021fairness} is to implicitly mitigate bias by adjusting the distribution of latent representations during the model training.

\textbf{Correlation Assumptions.} 
In the real-world scenario, it is hard to get the true distribution of sensitive attributes due to privacy concerns, we thus assume that the unfair model predictions are caused by certain related attributes that have high correlations to sensitive attributes.
Specifically, when we face the fairness issue for model prediction, it is challenging to leverage the model bias if we lack sensitive feature information.
Thus, there are some works that focus on eliminating prediction bias under the constraint of unknown sensitive attributes' distribution.
ARL~\cite{lahoti2020fairness} utilizes adversarial learning based on Rawlsian Max-Min fairness objectives. However, this approach could be too strict in enhancing fairness across groups, and it is hard to maintain the performance of downstream tasks.
FairRF~\cite{zhao2022towards} addresses the biased issues by leveraging the relatedness between a set of related non-sensitive attributes and sensitive attributes. 
This work assumes that the bias of model prediction actually comes from the high correlation between non-sensitive attributes and sensitive features. 
In this manner, a fair model can be achieved by the proposed objective function of alleviating the relatedness between non-sensitive attributes and sensitive attributes. Formally, the objective function of FairRF can be illustrated as follows:
Let $f_i \in F_n$ be a set of predefined related non-sensitive attributes, where $F_n$ is a set of non-sensitive features, FairRF applies correlation regularization $R_{related}$ on each $f_i$ to make trained model fair toward sensitive attribute $s$ by calculating the following function:
\begin{align}
    \min_{\mathcal{\theta}} \mathcal{R}_{\text{related}} = \sum_{i=1}^{K} \lambda_i \cdot \mathcal{R}(f_i, \hat{y}),
\end{align}
where $\lambda$ is the weight for regularizing correlation coefficient between $x^i$ and $\hat{y}$.

However, this correlation assumption between sensitive and non-sensitive attributes may be sub-optimal, because it requires strong assumptions on feature dependencies.
In other words, data-specific and distribution similarity are necessary.
For example, when we define the related features of sensitive features \textit{Gender} are three non-sensitive features, which are \textit{Age}, \textit{Relation}, and \textit{Marital-Status}, an accompanying assumption comes up that the three non-sensitive features have top-3 highest correlation with the given sensitive feature \textit{Gender}.
Nevertheless, it is possible that the true highest correlation-related features of the sensitive feature in the dataset are not obvious for human beings, so we cannot define it correctly.
For instance, maybe the highest related features of the sensitive feature \textit{gender} are \textit{color of eyes} and \textit{sleeping quality} in a certain dataset, and it is hard for humans to associate the two features as related features of \textit{gender}.
In our work, instead of adopting assumptions on bias feature distribution with its related features, we propose an assumption-free framework for automatically detecting the related features for bias mitigation.

\subsection{Learning Feature Interactions}
One major advantage of neural networks is their ability to model complex interactions between features by automatic feature learning. 
In the territory of click-through rate prediction, CTR prediction, feature interaction modeling has been playing a key role in improving downstream task performances by modeling different orders of feature combinations.
Instead of multiple layers of non-linear neural network approaches which suffer from inefficient and lack of good explanation of feature interactions~\cite{beutel2018latent}, there are popular approaches that are able to explicitly model different orders of feature combinations and meanwhile offer good model interpretability.
One of the previous works models feature interactions by calculating the inner products between a feature embedding and a trainable matrix, afterward calculating the Hadamard product of another feature embedding~\cite{huang2019fibinet}.
AutoInt~\cite{song2019autoint} models feature interactions by adopting the key-value attention mechanism and using the conducted attention weights between all feature pairs to weighted-sum the all input feature embedding.
AutoInt utilizes the inner product operator $\psi(\cdot, \cdot)$ to define the similarity between two feature embeddings $e_j$ and $e_c$, and leverages it to compute the attention weights under a specific attention head $h$ by the following equation:
\begin{align}
    a^{(h)}_{j, c} = \frac{exp(\psi^{(h)}(e_j, e_c))}{\sum_{n=1}^{N} {exp(\psi^{(h)}(e_j, e_n))}},
\end{align}
where $N$ represents the number of input features.

The classic self-attention-based approach considers all feature pairs for feature interaction learning, therefore it is difficult to significantly identify the bias between feature pairs containing target sensitive attributes.
In our work, we only consider the feature pairs which treat target sensitive attributes as a $Query$ of attention components to identify the feature interactions between sensitive and non-sensitive attributes for further alleviating the biased interactions.
Our framework can automatically detect the related features for bias mitigation.

\subsection{Problem Definition}
\begin{sloppypar}
We first define the notations used in this work. 
Let $\textbf{X}$ be the input data set and $\textbf{Y}$ be the ground truth label set of the model output, where $\textbf{X} = \{ x_1, \dots, x_{p}\}$ is the $p$-kind attribute set and $\textbf{Y} \in \{0, 1\}$ is the binary label set.
Among the input attribute set $\textbf{X} = \textbf{S} \cup \textbf{C}$, where sensitive attributes set $\textbf{S}$ (e.g. gender, race, marital status) and non-sensitive attributes set $\textbf{C}$.
We observe that the biased feature interactions are the influential factor in yielding fairness of predictive results. 
Formally, we define the sensitive feature interaction set as $\mathcal{I}_s = \left\{ \mathcal{I}(s, c_1), \dots , \mathcal{I}(s, c_{p-1}) |~ \forall c_j \in \textbf{C} \right\}$, where $\mathcal{I}(\cdot, \cdot)$ denotes an feature interaction between any two features, and $s \in \textbf{S}$ is a sensitive attribute.
For example, an interaction between a sensitive attribute \textit{gender} and non-sensitive attribute \textit{job} can be denoted as $\mathcal{I}(\textit{gender}, \textit{job})$. 
Based on modeling the feature interactions throughout the prediction models, the biased interactions from $\mathcal{I}_s$ eventually lead to bias on prediction tasks.

Based on the definitions and the intuitions above, we consider the interaction bias from prediction model $f(\textbf{X}, \theta) \equiv  p(g(\textbf{X}))$, where $\theta$ is the model parameters and $p(\cdot)$ is a single-layer prediction head of $d$-dimensional feature embedding encoder $g(\cdot): \textbf{X} \rightarrow \mathbb{R}^d$.
In our work, let $\mathcal{I}_s$ be the sensitive feature interaction set learned from prediction model $f(\cdot)$, we aim to identify the biased interaction that appears in $\mathcal{I}_s$ such that the detected biased interactions are alleviated during the prediction model training.
\end{sloppypar}

\begin{figure*}[ht]
\begin{center}
    \includegraphics[width=1\textwidth]{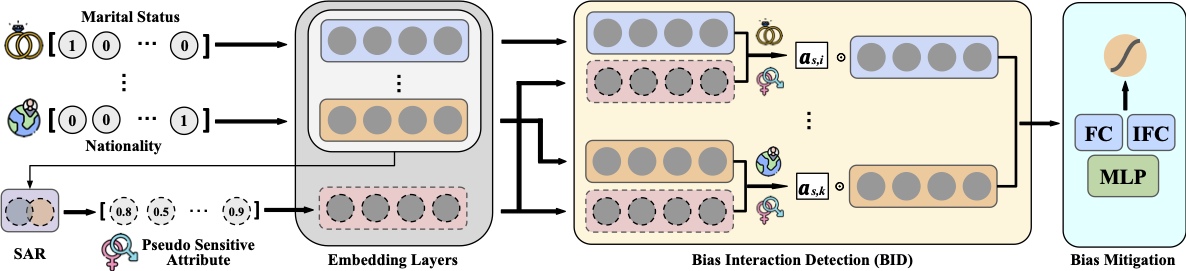}
    \caption{Overview of proposed FairInt framework. The FairInt framework includes three components, which are \textit{Sensitive Attributes Reconstructor (SAR)} to simulate the pseudo sensitive attributes and advantage the framework to better capture the interactions between the unavailable sensitive and non-sensitive attributes, \textit{Bias Interaction Detection (BID)} to identify the potential biased interactions, and \textit{Bias Mitigation} to alleviate the identified biased interactions by adopting two fairness regularizations \textit{Fairniess Constraint (FC)} and \textit{Interaction Fairness Constraint (IFC)}. Note that $a_{\hat{s}, c}$ is the attention weight between pseudo sensitive and non-sensitive attributes to weighted-sum all the feature embeddings, which we will describe in Sec.~\ref{sec:biasdet}.}
\label{fig:model_layout} 
\end{center}
\vspace{-0.2cm}
\end{figure*}

\section{Methodology}
In this section, we introduce an assumption-free fair mitigation framework, \textit{FairInt}, to alleviate the biased feature interactions.
Figure~\ref{fig:model_layout} illustrates our FairInt framework with two components: Assumption-free Bias Detection, which includes Sensitive Attributes Reconstructor (SAR) and Bias Interaction Detection (BID) layer, and Interaction-wise Bias Mitigation, which includes the regularizations Fairness Constraint (FC) and Interaction Fairness Constraint (IFC).
Our goal is to encourage the classifier to disentangle the biased interaction between sensitive and non-sensitive attributes and instead focus more on learning task-relevant information.
Assumption-free Bias Detection aims at identifying bias within feature interactions without predefined related features, and Interaction-wise Bias Mitigation focuses on alleviating the identified feature interaction bias.
In the following sections, we give a comprehensive description of our FairInt framework.
We first illustrate the details of the proposed bias detection component (Sec.~\ref{sec:biasdet}).
Then, we introduce our two bias mitigation components (Sec.~\ref{sec:biasmit}).
Finally, we demonstrate how to learn the fair predictor through our FairInt framework (Sec.~\ref{sec:fairint}).

\subsection{Assumption-free Bias Detection}\label{sec:biasdet}




sensitive attributes $s \in \textbf{S}$ are generally unavailable in real-world scenario during the inference stage. Many existing works mitigate the interaction bias under the assumption of the known distribution of sensitive attributes. However, in real-world scenarios, the unavailability of sensitive attributes exists due to various reasons, such as legal issues, which make most of the existing advancements unworkable. To tackle the problems, we develop two corresponding components: Sensitive Attributes Reconstructor (SAR) for sensitive attributes bias assumption-free, and Bias Interaction Detection (BID) for feature interaction assumption-free. Our assumption-free framework aims to disentangle the hand-crafted assumptions of the feature dependency between sensitive and specific non-sensitive attributes during the debiasing process.

\textbf{Sensitive Attributes Reconstructor (SAR).} 
Since sensitive attributes $s \in \textbf{S}$ are generally unavailable in real-world scenario during the inference stage, we design Sensitive Attributes Reconstructor (SAR) to simulate the sensitive attributes for alleviating the implicit interaction bias obtaining in non-sensitive attributes.
Specifically, we aim to generate a pseudo-sensitive attribute $\hat{s}$ by imitating the distribution of sensitive attributes $s \in \textbf{S}$ throughout our proposed reconstructor, which brings out the biased interaction between the sensitive attributes and all other non-sensitive features.

Let the input attribute set be $x \in \textbf{X}$ without the sensitive attributes $s \in \textbf{S}$. The objective of Sensitive Attributes Reconstructor (SAR) is to construct a reconstructor $f$ to generate a pseudo-sensitive attribute $\hat{s}$ for identifying the implicitly biased interactions toward non-sensitive features. The generating process of a pseudo-sensitive attribute can be formally illustrated as follows:
\begin{align}
    \hat{s} = \text{SAR}(\textbf{e}_{x/s}; \Theta_{r}),
\end{align}
where $\Theta_{r}$ is the trainable parameters of reconstructor $r$, and $\textbf{e}_{x/s}$ denotes the latent representation set of input features $x$ without sensitive attribute $s$.
Specifically, we leverage the embeddings of all non-sensitive attributions to generate a pseudo-sensitive attribute vector. This makes the reconstructor extract the correlated information between sensitive and non-sensitive features. During  training stage, the reconstructor loss $\mathcal{L}_{SAR}$ can be shown as follows:
\begin{align}
    \mathcal{L}_{SAR} \equiv \arg \min_{\Theta_{r}} \sum_{i=1}^{N} (\hat{s}_i - s_i)^2,
\label{func:sar}
\end{align}
where $N$ is the number of training instance.

The effectiveness of SAR was evaluated by predicting unavailable sensitive attributes using non-sensitive features from Adult and Law School datasets. SAR achieved 87\% accuracy for predicting \textit{Sex} in Adult and 94\% for predicting \textit{Race} in Law School.
The results show that SAR can achieve impressive performance by capturing the correlations between non-sensitive attributes and unobserved sensitive attributes.

Besides predicting the pseudo-sensitive attributes $\hat{s}$, SAR advantages our \textit{FairInt} to better capture the interactions between unobserved sensitive and non-sensitive attributes.

\textbf{Bias Interaction Detection (BID) Layer.}
Optimizing Eq. ~\ref{func:sar} in SAR generates a pseudo-sensitive attribute $\hat{s}$ as a sensitive sensitive attribute, which allows our proposed FairInt to quantitatively analyze the interaction between pseudo-sensitive attributes and non-sensitive attributes. Thus, we propose Bias Interaction Detection (BID) to identify the highly potential biased interactions with the generated pseudo-sensitive attribute.
We first let all the input features be the p-kind attribute set $\textbf{X} = \{x_1, \dots, x_p\}$ which contains categorical and numerical features.
Because categorical features are too sparse to learn, we map all the input features into low-dimensional spaces with the unique feature embeddings $e_i$. The formula can be illustrated as $e_i = M_i x_i$,
where $M_i$ is an embedding lookup matrix corresponding to feature $x_i$ with dimension $d$.

Feature interactions are typically modeled by either the inner product similarity or attention scoring mechanism between two feature embeddings~\cite{huang2019fibinet, song2019autoint}. For instance, AutoInt~\cite{song2019autoint} utilizes the multi-head self-attention mechanism to model high-order feature interactions for improving the downstream task's performance.
AutoInt learns feature interaction within a hierarchical representation structure, which has proven to be effective in several machine learning territories~\cite{tsang2017detecting, tsang2020feature}.
Especially, self-attention-based mechanism has been utilized in several machine learning areas for capturing the importance within features of input instances~\cite{vaswani2017attention}.
In our work, we exploit self-attention machanism~\cite{vaswani2017attention} to model feature interactions. The main goal of our framework is to mitigate the biased feature interaction for the model predictions but without predefined assumptions.
Therefore, based on the ability of self-attention mechanism to identify important feature interactions, we design Bias Interaction Detection (BID) to point out the key biased interactions of pseudo-sensitive attributes.

Unlike original self-attention mechanism that calculates the attention weights between all the feature two by two, we focus on modeling the feature interactions only between the sensitive pseudo-sensitive attribute $\hat{s}$ and other non-sensitive features by computing their attention weights.
Specifically, we model the interactions between a pseudo-sensitive attribute $\hat{s}$ and one non-sensitive features $c \in \textbf{C}$ with attention head $h$ as $a_{\hat{s}, c}$, which can be calculated as follows:
\begin{align}
    a_{\hat{s}, c} = \frac{\exp(\psi^{h}(\hat{e_{s}}, e_c))}{\sum_{c \in \textbf{C}} {\exp(\psi^{h}(\hat{e_{s}}, e_c))}},
\label{eq:attweight}
\end{align}
where $\hat{e_{s}}$ and $e_c$ are the low-dimensional embedding of $\hat{s}$ and $c$, and $\psi^{h}(\hat{e_{s}}, e_c)$ denotes as the scoring operator to evaluate the similarity between $\hat{e_{s}}$ and $e_c$.
In this paper, we adopt dot product as an example for $\psi^{h}(\hat{e_{s}}, e_c)$, which can be illustrated as follows:
\begin{align}
    \psi^{h}(\hat{e_{s}}, e_c) = \langle W^{h}_{\text{Query}} \hat{e_{s}}, W^{h}_{\text{Key}} e_c \rangle,
\end{align}
where $\langle \cdot ~, \cdot \rangle$ is inner product operator, and $W^h_{\text{Query}}$ and $W^h_{\text{Key}}$ are embedding matrices for $\hat{e_{s}}$ and $e_c$. The biased interaction scores can now be defined as $a_{\hat{s}, c}$ in this manner.

After obtaining the biased interaction scores between the sensitive and non-sensitive features, we generate the biased interaction embeddings $\hat{e}^{H}_s$ to represent the biased interactions for bias mitigation. We formally define the biased interaction embeddings as following formula:
\begin{align}
    \hat{e}^{H}_s = \sometext_{h=1}^{|\textbf{H}|} ~ \sum_{c=1}^{C} a_{\hat{s}, c} (W^{h}_{value}\cdot e_c),
\label{eq:weightedsum}
\end{align}
where $W^{h}_{value}$ is a trainable embedding matrix, and $\Vert$ denotes the concatenation operator for all biased interaction embeddings of each attention layer $h \in \textbf{H}$.

\vspace{-0.1cm}
\subsection{Interaction-wise Bias Mitigation}\label{sec:biasmit}



After receiving the detected bias interaction embeddings $\hat{e}^{H}_s$, we focus on alleviating the bias from feature interactions.
Our goal is to equalize the conditional probability distribution of bias interaction embeddings given different sensitive attributes $s \in \textbf{S}$. However, the sensitive attribute information in $\hat{e}^{H}_s$ can be easily perturbed due to the imbalance amounts of sensitive and non-sensitive attributes. This may affect the bias mitigation performance since the alleviation process requires an explicate sensitive attribute as a pivot to mitigate. Hence, we adopt a residual neural network (ResNet)~\cite{he2016deep, song2019autoint} to enrich the information of pseudo-sensitive attributes, which we can formally reveal as follows:
\begin{align}
    \overline{e}_{\hat{s}} = ReLU(\hat{e}^{H}_s + W_{Res}\cdot \hat{e_{s}}),
\label{eq:resnet}
\end{align}
where $W_{Res}$ is the residual model weight and $\hat{e_{s}}$ is the embedding of pseudo-sensitive attributes.
In this work, we design two fairness constraints: Interaction Fairness Constraint and Fairness Constraint for biased interaction mitigation.

\textbf{Interaction Fairness Constraint (IFC) Loss.}
In order to mitigate the detected bias interactions from different sensitive attribute groups, we design the Interaction Fairness Constraint (IFC) loss to minimize the KL-divergence between the sensitive attribute groups. IFC can then ensure the equivalent information gained from each feature interaction.
Formally, IFC can be formulated as follows:
\begin{align}
    \mathcal{L}_{IFC} = \sum_{i \in \textbf{S}} \sum_{j \in \textbf{S}/i} \text{KL}(\overline{e}_{[\hat{s} \approx i]}, \overline{e}_{[\hat{s} \approx j]}),
\label{eq:ifc}
\end{align}
where $\text{KL}(\cdot)$ denotes the KL-Divergence, and $\overline{e}_{[\hat{s}\approx i]}$ is the subset of $\overline{e}_{\hat{s}}$ that is more similar to sensitive attributes $i \in \textbf{S}$. To the convenience of our work, we set the hierarchical boundary with expected value of uniform distributed $\textbf{S}$ to distinguish which group $\hat{s}$ belongs to in $\textbf{S}$.
IFC loss mitigates the bias information of the latent representation by calculating the KL-divergence scores as biased scores between each group in pseudo-sensitive attributes $\textbf{S}$.
Therefore, by adding $\mathcal{L}_{IFC}$ as a regularization term to our framework, the bias feature interaction of latent representation $\overline{e}_{\hat{s}}$ can be alleviated.

\textbf{Fairness Constraint (FC) Loss.} 
Although our proposed IFC mitigates most of the biased interaction information from the embedding aspect, the remaining biased interaction may be amplified by prediction models and generate unfair task predictions.
To alleviate the unfairness of model predictions on downstream tasks, we adopt the Fairness Constraint (FC) loss toward pseudo-sensitive attributes $\hat{s}$. In this work, we focus on classification tasks.
Our proposed FC aims to mitigate biased prediction behaviors $\hat{y}$ by computing the absolute differences of the cross entropy between every two of each pseudo-sensitive attribute $(\hat{s}_i, \hat{s}_j) \in \textbf{S}$.
Formally, FC can be formulated as follows:
\begin{align}
    \mathcal{L}_{FC} = \sum_{i \in \textbf{S}} \sum_{j \in \textbf{S}/i} |CE_{[\hat{s} \approx i]} - CE_{[\hat{s} \approx j]}|,
\label{eq:fc}
\end{align}
where $CE_{[\hat{s} \approx i]}$ is cross entropy which belongs to a certain sensitive attribute $i \in \textbf{S}$.
Based on the meaning of cross entropy, it reflects the correctness of classification prediction $\hat{y}$. The idea of FC loss is to ensure the discrepancy of the correctness of $\hat{y}$ by giving every two of each pseudo-sensitive attribute.
Thus, $\mathcal{L}_{FC}$ can effectively alleviate the prejudiced model predictions among the pseudo-sensitive attribute set.

\vspace{-0.1cm}
\subsection{Fair Classifier with FairInt}\label{sec:fairint}
Here we discuss how to incorporate the IFC loss $\mathcal{L}_{IFC}$ and the FC loss $\mathcal{L}_{FC}$ with a classifier to alleviate the biased interaction.
We adopt the reconstructor loss $\mathcal{L}_{SAR}$ to our framework for training the SAR to generate the pseudo-sensitive features.
As the IFC loss can be a stand-alone optimization objective function, it is capable of mitigating bias feature interaction in latent representations for any kinds of classification model.
In our work, we evaluate the effectiveness of our framework on a one-layer multi-layer perceptron as the classification model, which can be replaced by any deeper or more powerful models.

To train a classification model with our proposed FairInt framework, we optimize the cross entropy loss $\mathcal{L}_{0}$.
We then incorporate $\mathcal{L}_{0}$ with the Interaction Fairness Constraint (IFC) loss $\mathcal{L}_{IFC}$, Fairness Constraint (FC) loss $\mathcal{L}_{FC}$, and reconstructor loss $\mathcal{L}_{SAR}$ as the final objective function to fair classifier training.
Our proposed IFC loss and FC loss help the classification models mitigate the bias feature interactions from the views of latent representations and alleviate the prejudiced model predictions with given different kinds of sensitive attributes during training.
Specifically, we optimize the proposed FairInt by illustrating the following joint loss function:
\begin{align}
    \mathcal{L}_{\text{FairInt}} = \mathcal{L}_{0} + \lambda_{IFC} \mathcal{L}_{IFC} + \lambda_{FC} \mathcal{L}_{FC} + \mathcal{L}_{SAR},
\label{eq:lossfunc}
\end{align}
where $\mathcal{L}_{\text{FairInt}}$ denotes as the loss function to the proposed FairInt and $\lambda_{IFC}$ and $\lambda_{FC}$ are the weighting hyper-parameters to balance the biased interaction mitigating and feature interactions modeling.
By optimizing $\mathcal{L}_{\text{FairInt}}$, we can alleviate the bias model predictions by mitigating the detected bias feature interactions without defining any related and potentially biased features interactions in advance.
In inference stage, the trained FairInt framework can provide fair predictions without sensitive attributes.

\section{Experiment}\label{sec:exp}
In this section, we empirically evaluate the proposed \textit{FairInt} framework. We mainly focus on the following research questions:
\begin{enumerate}[label=\textbf{Q\arabic*},leftmargin=*]
	 \item \label{q1} Compared with the existing baseline methods, can our assumption-free \textit{FairInt} framework mitigate the unfair model prediction on the downstream tasks (Sec.~\ref{sec:perf})?
	 \item \label{q2} Can our proposed Bias Interaction Detection layer identify the bias feature interaction and encode it in the latent representation (Sec.~\ref{sec:analybid})?
	 \item \label{q3} How do the proposed Interaction Fairness Constraint loss and Fairness Constraint loss in Eq.~\ref{eq:ifc} and Eq.~\ref{eq:fc} impact the fairness of the classification model (Sec.~\ref{sec:analyfc})?
	 \item \label{q4} How do the hyper-parameters impact the fairness performance of the proposed \textit{FairInt} (Sec.~\ref{sec:analyhp})?
	 \item \label{q5} How does our assumption-free \textit{FairInt} framework automatically detect the related features and further mitigate the bias feature interaction (Sec.~\ref{sec:observ})?
\end{enumerate}

\subsection{Datasets}

We consider four real-world tabular datasets~\cite{quy2021survey} that are commonly used for studying fairness-aware classification, which include three application domains as shown in Table~\ref{tab:ds}.

\begin{table}[t!]
\vspace{-0.2cm}
\caption{Dataset statistic.}
\vspace{-0.2cm}
\resizebox{0.45\textwidth}{!}{%
\begin{tabular}{@{}lllll@{}}

\toprule
                    & Adult   & Law School & Bank Marketing & Diabetes  \\ \midrule
\# Instances         & 48,842   & 20,798      & 45,211          & 101,766     \\
\# Attributes         & 15      & 12         & 17             & 50        \\
Data domain         & Finance & Education  & Finance        & Heathcare \\
Sensitive attribute~\cite{quy2021survey} & Gender  & Race       & Marital status & Gender  \\
\bottomrule
\end{tabular}%
}
\label{tab:ds}
\vspace{-0.5cm}
\end{table}

\begin{table*}[tbh!]
    \large
    \centering
    \caption{Performance of classification on the real-world datasets, Adult, Law School, Bank Marketing, Diabetes, where \textit{Vanilla} is an MLP model, \textit{FC} and \textit{PR} are two MLP models that adopted the two fairness regularizations. Recall that lower $\Delta DP$ and $\Delta EO$ are better while larger $AUC$ is better. 
    Because \textit{FairRF}~\cite{zhao2022towards} cannot perform well trade-off between AUC and fairness performance, we don't compare other method's $\Delta DP$ and $\Delta EO$ with \textit{FairRF}.
    The $\dagger$ denotes the comparison baseline of each metric for \textit{FairInt}, PR~\cite{kamishima2012fairness}. The improvements of \textit{FairInt} compared to the baselines are shown in the row \textit{Improv.}}
    \vspace{-0.2cm}
    \makebox[1\textwidth][c]{
        \resizebox{1\textwidth}{!}{
            \begin{tabular}{l|ccc|ccc|ccc|ccc}
                \toprule
                 & \multicolumn{3}{c}{Adult} |& \multicolumn{3}{c}{Law School} |& \multicolumn{3}{c}{Bank Marketing} |& \multicolumn{3}{c}{Diabetes} \\
                 \cmidrule(lr){2-4}\cmidrule(lr){5-7}\cmidrule(lr){8-10}\cmidrule(lr){11-13}
                 & AUC & $\Delta DP$ & $\Delta EO$ & AUC & $\Delta DP$ & $\Delta EO$ & AUC & $\Delta DP$ & $\Delta EO$ & AUC & $\Delta DP$ & $\Delta EO$ \\
                 \midrule
                 Vanilla & 0.9097 & 0.1864 & 0.1013 & 0.8921 & 0.0386 & 0.0187 & 0.9218 & 0.0360 & 0.0545 & 0.6107 & 0.0075 & 0.0085 \\
                 AutoInt~\cite{song2019autoint} & \textbf{0.9108} & 0.2038 & 0.1027 & \textbf{0.8934} & 0.0441 & 0.0247 & \textbf{0.9251} & 0.0389 & 0.0754 & \textbf{0.6152} & 0.0057 & 0.0115 \\
                 \midrule
                 FC & 0.9065 & 0.1729 & 0.0552 & 0.8790 & 0.0146 & 0.0061 & 0.9178 & 0.0198 & 0.0343 & 0.6086 & 0.0051 & 0.0063 \\
                 PR $\dagger$~\cite{kamishima2012fairness} & 0.8862 & 0.1676 & 0.0880 & 0.8804 & 0.0121 & 0.0089 & 0.9091 & 0.0238 & 0.0303 & 0.6037 & 0.0057 & 0.0057 \\
                 ARL~\cite{lahoti2020fairness} & 0.8807 & 0.1107 & 0.0608 & 0.8310 & 0.0091 & 0.0146 & 0.8309 & 0.0099 & 0.0199 & 0.6021 & 0.0039 & 0.0044 \\
                 FairRF~\cite{zhao2022towards} & 0.8358 & 0.0343 & 0.0797 & 0.7622 & 0.0080 & 0.0176 & 0.8158 & 0.0120 & 0.0280 & 0.6031 & 0.0031 & 0.0038 \\
                 \midrule\midrule
                 FairInt & 0.8826 & \textbf{0.1586} & \textbf{0.0525} & 0.8829 & \textbf{0.0077} & \textbf{0.0038} & 0.9140 & \textbf{0.0182} & \textbf{0.0249} & 0.6064 & \textbf{0.0041} & \textbf{0.0034} \\
                 
                 Improv. & -0.41\% & \textbf{-5.37\%} & \textbf{-40.34\%} & +0.28\% & \textbf{-36.36\%} & \textbf{-57.30\%} & +0.54\% & \textbf{-23.53\%} & \textbf{-17.82\%} & +0.45\% & \textbf{-28.07\%} & \textbf{-40.35\%} \\
                 \bottomrule
            \end{tabular}%
        }
    }
    \label{tab:tb1}
\end{table*}%

\begin{figure*}[tbh!]
    \begin{subfigure}{.245\textwidth}
      \centering
      \includegraphics[width=1\linewidth, trim=0 0 0 0,clip]{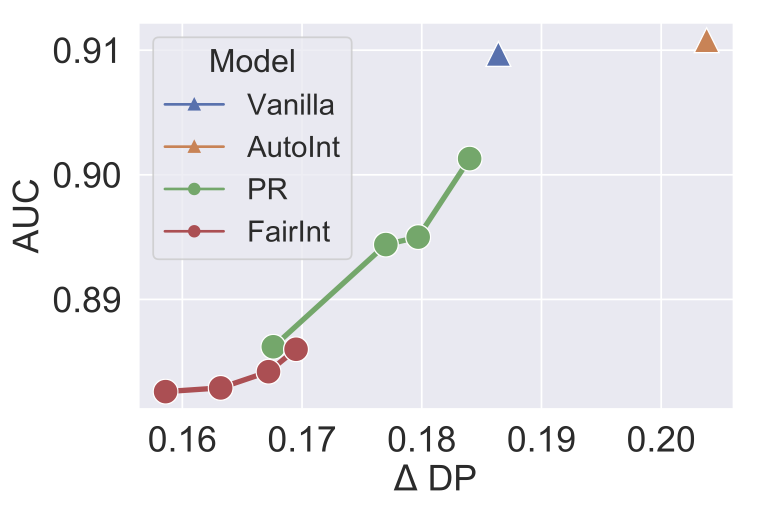}
    \end{subfigure}
    \begin{subfigure}{.245\textwidth}
      \centering
      \includegraphics[width=1\linewidth, trim=0 0 0 0,clip]{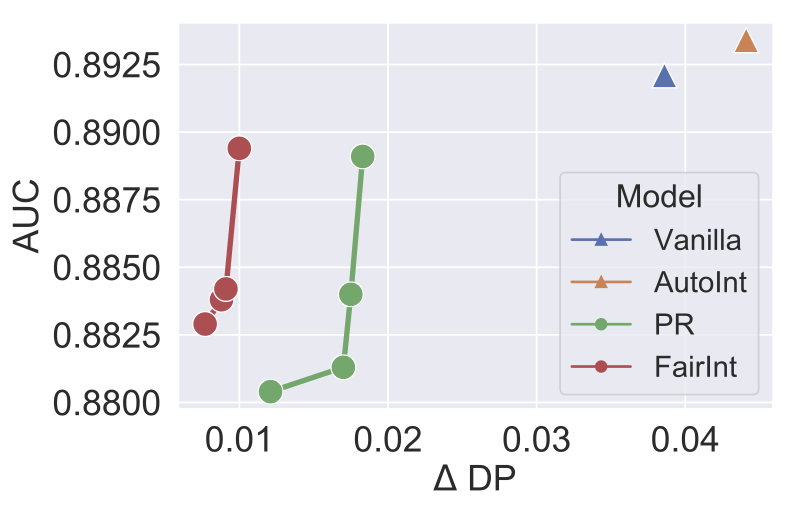}
    \end{subfigure}
    \begin{subfigure}{.245\textwidth}
      \centering
      \includegraphics[width=1\linewidth, trim=0 0 0 0,clip]{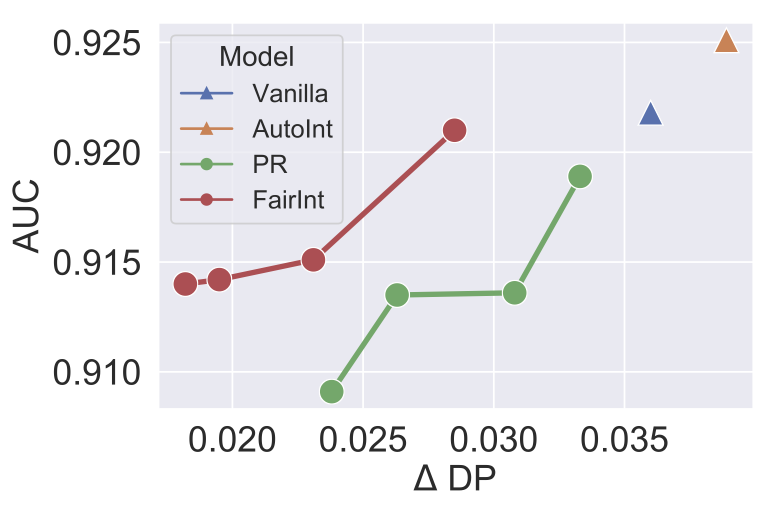}
    \end{subfigure}
    \begin{subfigure}{.245\textwidth}
      \centering
      \includegraphics[width=1\linewidth, trim=0 0 0 0,clip]{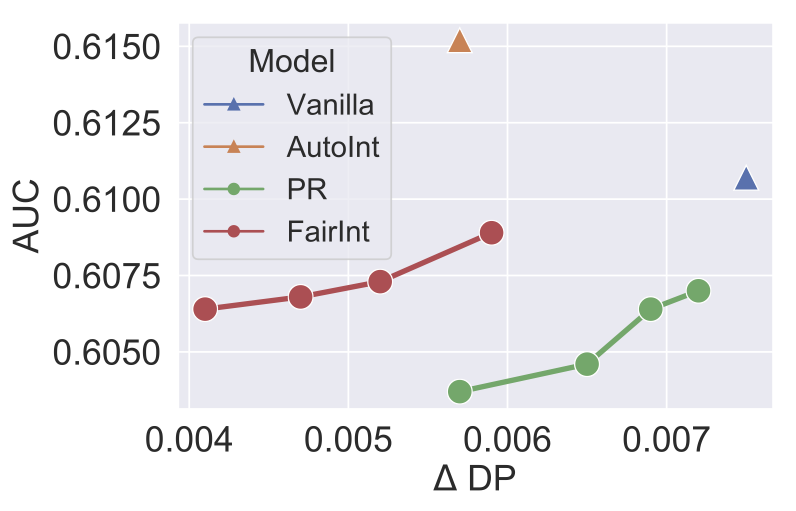}
    \end{subfigure}
    \begin{subfigure}{.245\textwidth}
      \centering
      \includegraphics[width=1\linewidth, trim=0 0 0 0,clip]{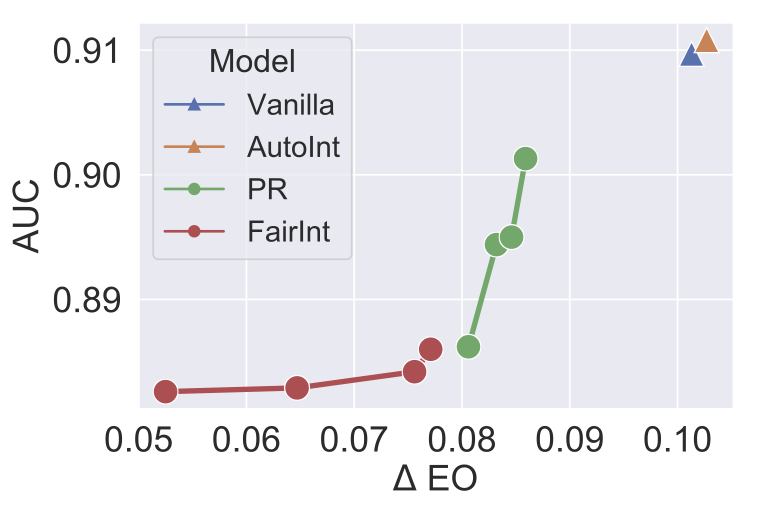}
      \subcaption{Adult}
    \end{subfigure}
    \begin{subfigure}{.245\textwidth}
      \centering
      \includegraphics[width=1\linewidth, trim=0 0 0 0,clip]{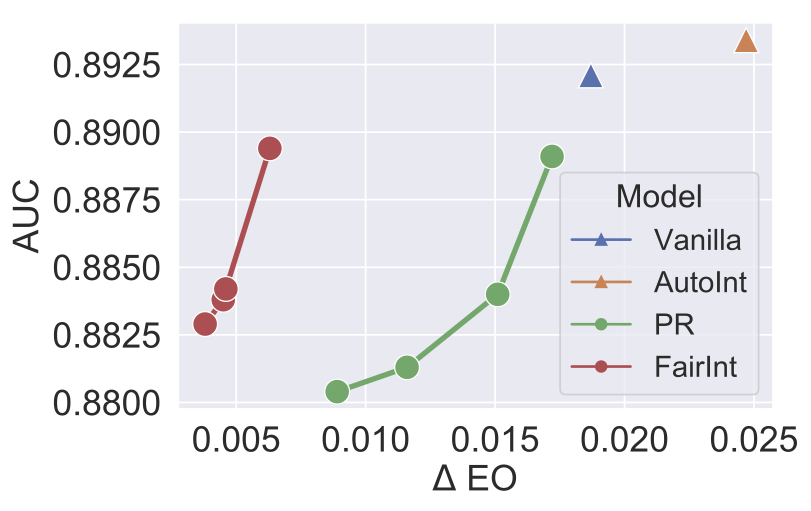}
      \subcaption{Law School}
    \end{subfigure}
    \begin{subfigure}{.245\textwidth}
      \centering
      \includegraphics[width=1\linewidth, trim=0 0 0 0,clip]{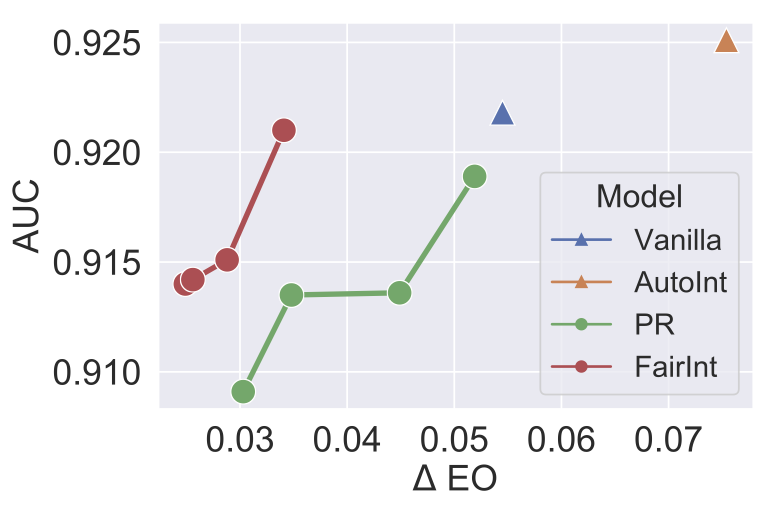}
      \subcaption{Bank Marketing}
    \end{subfigure}
    \begin{subfigure}{.245\textwidth}
      \centering
      \includegraphics[width=1\linewidth, trim=0 0 0 0,clip]{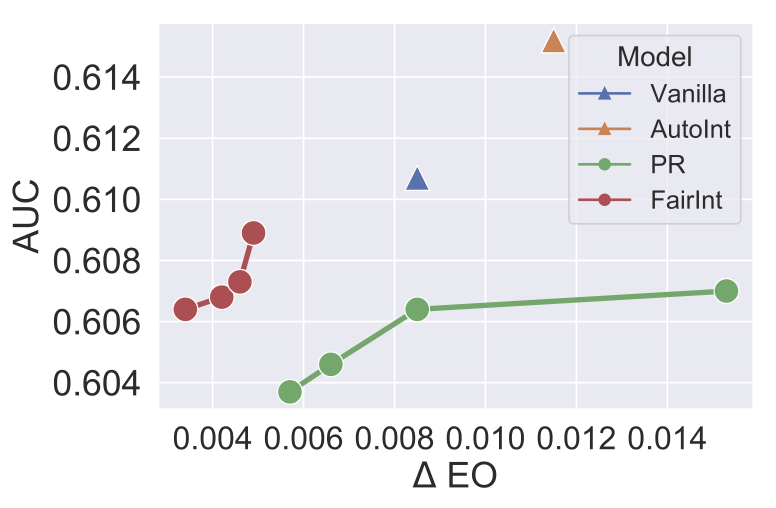}
      \subcaption{Diabetes}
    \end{subfigure}
    \vspace{-0.2cm}
    \caption{The fairness-AUC curve comparison of FairInt and other baselines, where \textit{PR} denotes an MLP model that adopted the fairness regularization \textit{Prejudice Remover (PR)}. Recall that lower $\Delta DP$ and $\Delta EO$ are better while larger $AUC$ is better. The first and second rows represent the $\Delta DP$ and $\Delta EO$ trade-off curves, respectively.}
    \vspace{-0.2cm}
    \label{fig:trade-off}
\end{figure*}

\subsection{Baselines and Fairness Metrics}
Besides the \textit{ARL}~\cite{lahoti2020fairness} and \textit{FairRF}~\cite{zhao2022towards} mentioned in Sec~\ref{sec:bias_mit}, we also leverage two fairness constraint regularization methods to train vanilla MLP models as baselines for comparing with our framework.
We adopt the Fair Constraint (FC) loss as a regularization to a vanilla MLP model as a baseline, where the FC loss is to mitigate biased behaviors toward model predictions, and it can be calculated as Eq.~\ref{eq:fc}.
For another baseline, we apply a regularization-based mitigation method to a vanilla MLP model, Prejudice Remover~\cite{kamishima2012fairness}, which considers the mutual information for equalizing the distributions between two variables to alleviate biases.
For each dataset, both the two baselines are leveraged to the same vanilla MLP model which we will describe in the Sec.~\ref{sec:impl}.
We also compare the proposed \textit{FairInt} to two other baselines including vanilla MLP classification models and the CTR prediction model AutoInt, which modeling feature interaction by adopting the key-value attention mechanism to improve the performance on CTR prediction tasks.

We use two group fairness metrics to evaluate the fairness of prediction models: Demographic Parity ($\Delta DP$)~\cite{wan2021modeling} and Equalized Odds ($\Delta EO$)~\cite{wan2021modeling}.
$\Delta DP$ measures the difference in the probability of a positive outcome between different sensitive groups and it is better to be closer to 0, which can be calculated as follows:
\begin{align}
    \Delta DP = p(\hat{y} = 1|s = s_i) - p(\hat{y} = 1|s = s_j),
\label{eq:dp}
\end{align}
where $s_i$ and $s_j$ represent different sensitive groups.
Equalized Odds require the probability of positive outcomes to be independent of the sensitive group $s$, conditioned on the ground truth label $y$.
Specifically, $\Delta EO$ calculates the summation of the True Positive Rate difference and False Positive Rate difference as follows:
\begin{multline}
    \Delta EO = |P(\hat{y} = 1|s = s_i, y = 1) - P(\hat{y} = 1|s = s_j, y = 1)| \\ 
    + |P(\hat{y} = 1|s = s_i, y = 0) - P(\hat{y} = 1|s = s_j, y = 0)|,
\end{multline}
\label{eq:eo}
where $\Delta EO$ is better to be closer to 0.

\subsection{Implementation Details}\label{sec:impl}
In \textit{FairInt}, we set the embedding dimension $d=4$ and the number of attention heads in BID layer as 1 among all four datasets.
For the Adult dataset, we use a two-layer MLP with 64 and 32 units of each hidden layer as the vanilla MLP model. 
For the Law School dataset, we use a two-layer MLP with 64 and 32 units of each hidden layer as the vanilla MLP model.
For the Bank Marketing dataset, we use a two-layer MLP with 40 and 20 units of each hidden layer as the vanilla MLP model.
For the Diabetes dataset, we use a four-layer MLP with 512, 256, 64, and 64 units of each hidden layer as the vanilla MLP model.
As for the AutoInt, we set the embedding dimension of each feature to 4, the number of interaction layers to 2, and the number of attention heads in each interaction layer to 2 for all four datasets.
To prevent overfitting, we search dropout rate from the set $\{0.1, 0.3, 0.5, 0.7, 1.0\}$ and search $l_2$ norm regularization from the set $\{\num{5e-1}, \num{1e-1}, \num{5e-2}, \num{1e-2}, \num{5e-3}, \num{1e-3}, \num{5e-4}, \num{1e-4}, \num{5e-5}, \num{1e-5}\}$ for all the models.
To address the situation of the  unavailablity of sensitive attributes in the inference stage, we utilize a four-layer MLP as the SAR on both vanilla MLP predictor and AutoInt models.

\subsection{(Q1) Performance Comparison on Real-world Datasets}\label{sec:perf}
In this section, we provide the results of classification prediction by using a binary classifier performance measure AUC~\cite{hanley1982meaning} for evaluating imbalanced data.
The fairness testings are evaluated with the two aforementioned fairness metrics: $\Delta DP$ and $\Delta EO$.
Table~\ref{tab:tb1} summarizes the performance of the \textit{FairInt} and the baselines on the four real-world datasets, where FC and PR refer to two vanilla MLP models which are debiased by two regularization-based bias alleviation methods Fair Constraint proposed in Sec.~\ref{sec:biasdet} and Prejudice Remover~\cite{kamishima2012fairness}.
We observe that our \textit{FairInt} significantly and consistently mitigates the bias predictions with the lower $\Delta DP$ and $\Delta EO$ across all four datasets.
The best fairness results are highlighted in bold.
Given the limitations demonstrated by \textit{FairRF}~\cite{zhao2022towards} and \textit{ARL}~\cite{lahoti2020fairness} in balancing the trade-off between AUC and fairness performance as assessed in the Law School and Bank Marketing, a comparison of their DP and EO performance with other methods is not performed.
Compared with the best bias alleviated baselines between FC and PR, our \textit{FairInt} improves $\Delta DP$ by 5.37\%, 36.36\%, 8.08\%, and 19.61\% on Adult, Law School, Bank Marketing, and Diabetes, respectively.
As for $\Delta EO$, our \textit{FairInt} improves $\Delta EO$ by 4.89\%, 37.70\%, 17.82\% and 40.35\% on the four datasets, respectively.
We also make the following observations of the experimental results.

\begin{table*}[tbh!]
    \large
    \centering
    \caption{Ablation study on the real-world datasets, Adult, Law School, Bank Marketing, Diabetes. The \textit{Vanilla FairInt} refers to the \textit{FairInt} framework without the two interaction-wise bias mitigation regularization IFC and FC, the \textit{+ IFC} and \textit{+ FC} denote the \textit{Vanilla FairInt} with the fairness regularizations \textit{IFC} and \textit{FC}, respectively. Compared with the complete \textit{FairInt}, the component FC improves more on $\Delta DP$ while IFC more focusing on improving  $\Delta EO$.}
    \vspace{-0.2cm}
    \makebox[1\textwidth][c]{
        \resizebox{1\textwidth}{!}{
            \begin{tabular}{l|ccc|ccc|ccc|ccc}
                \toprule
                 & \multicolumn{3}{c}{Adult} |& \multicolumn{3}{c}{Law School} |& \multicolumn{3}{c}{Bank Marketing} |& \multicolumn{3}{c}{Diabetes} \\
                 \cmidrule(lr){2-4}\cmidrule(lr){5-7}\cmidrule(lr){8-10}\cmidrule(lr){11-13}
                 & AUC & $\Delta DP$ & $\Delta EO$ & AUC & $\Delta DP$ & $\Delta EO$ & AUC & $\Delta DP$ & $\Delta EO$ & AUC & $\Delta DP$ & $\Delta EO$ \\
                 \midrule

                 Vanilla FairInt & \textbf{0.8885} & 0.1928 & 0.1347 & 0.8731 & 0.0448 & 0.0277 & 0.9142 & 0.0440 & 0.0973 & \textbf{0.6115} & 0.0104 & 0.0132 \\
                 \midrule
                 
                 + IFC & 0.8822 & 0.1658 & 0.0866 & 0.8746 & 0.0188 & 0.0056 & \textbf{0.9182} & 0.0286 & \textbf{0.0249} & 0.5927 & 0.0060 & 0.0055 \\
                 Improv. & --- & -14.00\% & -35.71\% & --- & -58.04 \% & -79.78\% & --- & -35.00\% & -74.41\% & --- & -42.31\% & -58.33\% \\
                 \midrule
                 
                + FC & 0.8770 & 0.1635 & 0.0638 & 0.8795 & 0.0139 & 0.0078 & 0.9127 & 0.0212 & 0.0262 & 0.5969 & 0.0055 & 0.0071 \\
                 Improv. & --- & -15.20\% & -52.64\% & --- & -68.97\% & -71.84\% & --- & -51.82\% & -73.07\% & --- & -47.12\% & -46.21\% \\
                 \midrule
                 
                 + IFC+ FC (FairInt) & 0.8826 & \textbf{0.1586} & \textbf{0.0525} & \textbf{0.8829} & \textbf{0.0077} & \textbf{0.0038} & 0.9140 & \textbf{0.0182} & \textbf{0.0249} & 0.6064 & \textbf{0.0041} & \textbf{0.0034} \\
                Improv. & --- & -17.74\% & -61.02\% & --- & -82.81\% & -86.28\% & --- & -58.64\% & -74.41\% & --- & -60.58\% & -74.24\% \\

                 \bottomrule
            \end{tabular}%
        }
    }
    \label{tab:tb2}
\end{table*}%

\begin{figure*}[tbh!]
\begin{center}
    \includegraphics[width=1\textwidth]{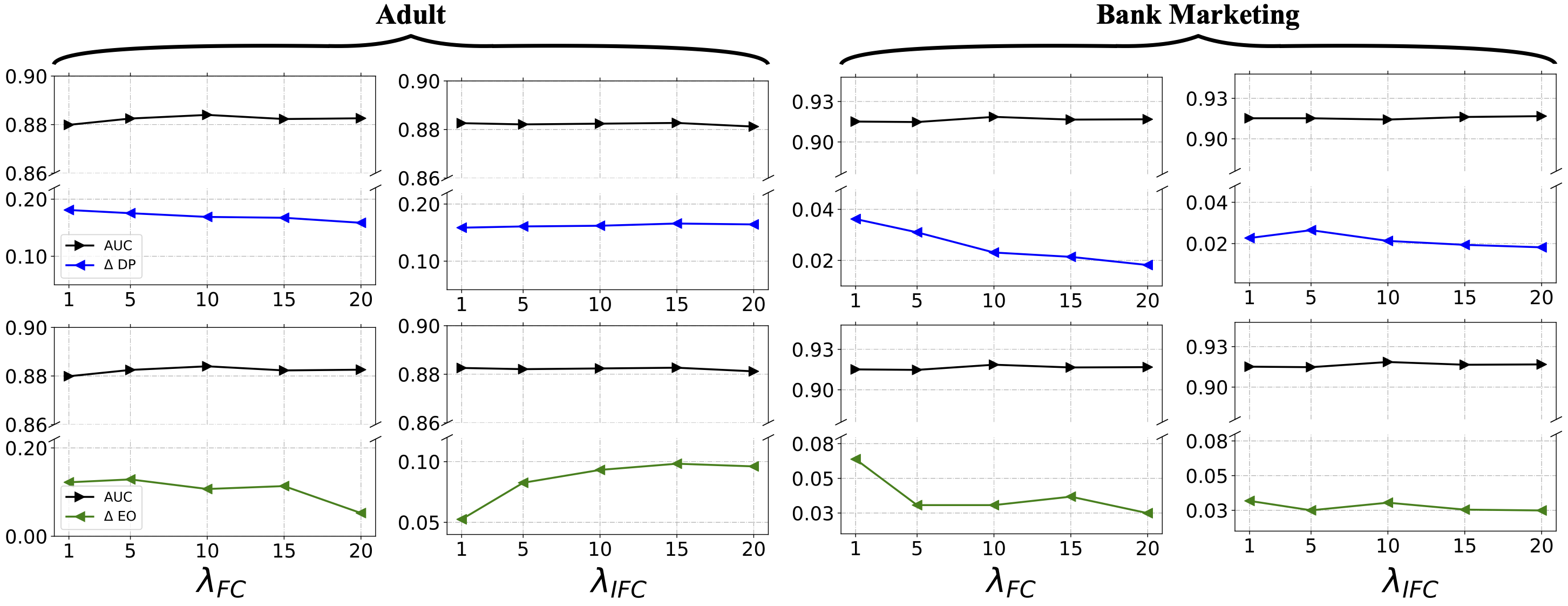}
    \vspace{-0.7cm}
    \caption{Sensitive analysis on $\lambda_{FC}$ and $\lambda_{IFC}$ for FairInt on Adult and Bank Marketing datasets. With the similar classification performance among different $\lambda_{FC}$, FairInt outperforms when $\lambda_{FC}$ is $20$ on fairness performance. As for $\lambda_{IFC}$, the best $\lambda_{IFC}$ can typically achieve the best $\Delta DP$ when reaching the best $\Delta EO$.}
\label{fig:sensi_fc_ifc}
\end{center}
\vspace{-0.2cm}
\end{figure*}

First, AutoInt can slightly improve the AUC performance with the attention-based feature interaction modeling mechanism, but it also augments the biased prediction behaviors.
As we can see from Table~\ref{tab:tb1}, AutoInt can improve the AUC of vanilla MLP models by 0.44\%, 0.15\%, 0.36\% and 0.74\% on the four datasets, respectively.
However, it at the same time increases $\Delta DP$ and $\Delta EO$ on all four datasets.
Compared with the vanilla MLP models, AutoInt increases $\Delta DP$ on three out of four datasets and increases $\Delta EO$ on all four datasets.
The reason is that the modeled feature interactions not only improve the downstream task performances but also contain the biased feature interactions that will augment the biased behaviors of predictions.

Second, our \textit{FairInt} can maintain the competitive classification performance compared with the other debiased baselines.
As we can see from Table~\ref{tab:tb1}, the fairness performances of our proposed \textit{FairInt} are improved significantly while the classification performances are slightly decreased.
We compare our \textit{FairInt} with the Vanilla MLP model, AutoInt, and the debiased baselines PR in Figure~\ref{fig:trade-off}, which illustrates their fairness-AUC curves for the four datasets.
The hyper-parameter $\lambda_{IFC}$ and $\lambda_{FC}$ in Eq.~\ref{eq:lossfunc} controls the trade-off between AUC and fairness for \textit{FairInt}.
For the debiased vanilla MLP with PR, the hyper-parameter in front of the regularization term also controls the trade-off.
From Figure~\ref{fig:trade-off} we also can observe that our proposed \textit{FairInt} can achieve the best $\Delta DP$ and $\Delta EO$ in all four datasets while remaining competitive AUC compared to PR.

\subsection{(Q2) Analysis of Bias Interaction Detection Layer}\label{sec:analybid}
We analyze the ability of the Bias Interaction Detection (BID) layer that can identify the biased feature interactions.
In Table~\ref{tab:tb2}, the \textit{Vanilla FairInt} refers to the \textit{FairInt} framework without the two interaction-wise bias mitigation regularization IFC and FC, and it keeps the Bias Interaction Detection (BID) layer which is designed to identify biased feature interactions.
Compared with the vanilla MLP models, the \textit{Vanilla FairInt} significantly augments biased behaviors of model predictions.
For all four datasets, the \textit{Vanilla FairInt} increases $\Delta DP$ by 2.99\%, 16.06\%, 22.22\% and 38.67\%, and it increases $\Delta EO$ by 33.10\%, 48.13\%, 37.24\% and 55.29\%, respectively.
The reason \textit{Vanilla FairInt} can remarkably augment the biased predictions is that BID focuses on detecting the biased feature interactions and embedding them into the latent representation.
A similar scenario can be observed in AutoInt because it models all the interactions between all the feature pairs that include biased feature interactions.
Unlike the AutoInt, our proposed \textit{FairInt} focuses on learning to model the biased feature interactions only among the feature pairs which contain a sensitive attribute.
By doing so, the latent representations in \textit{FairInt} embed the biased feature interaction information without other noising knowledge.

\begin{figure*}[tbh!]
\begin{center}
    \includegraphics[width=1\textwidth]{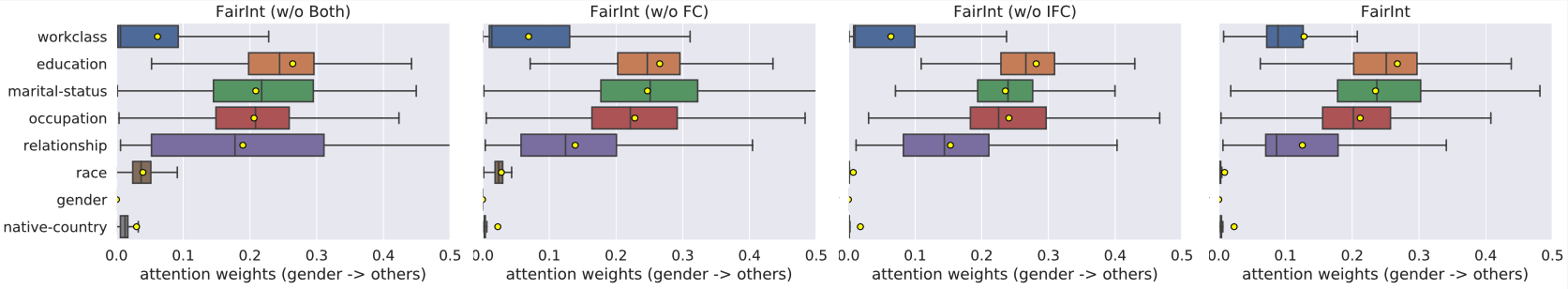}
    \caption{Observations of feature interaction in \textit{FairInt} on Adult datasets, where \textit{FairInt (w/o Both)} refers to the FairInt without the two fairness regularization component Fairness Constraint (FC) and Interaction Fairness Constraint (IFC), \textit{FairInt (w/o FC)} refers to the FairInt without FC, FairInt (w/o IFC) refers to the FairInt without IFC, and \textit{FairInt} refers to the complete FairInt. In the above four figures, the yellow points represent the mean values of each attention weight between the sensitive attribute \textit{gender} and a non-sensitive attribute. From the observations, we conclude that the variances of each interaction are lower after \textit{FairInt} achieved better fairness performance by adopting FC and IFC.}
\label{fig:att_wei}
\end{center}
\vspace{-0.2cm}
\end{figure*}
\vspace{-0.3cm}

\subsection{(Q3) Analysis of Fairness Constraint Components}\label{sec:analyfc}
After the latent representations in \textit{FairInt} embed the bias feature interaction information, we leverage the two fairness constraints to mitigate the embedded bias feature interactions.
To better understand the effects of the two fairness constraints, Interaction Fairness Constraint and Fairness Constraint, in the proposed \textit{FairInt}, we conduct the ablation studies to analyze and verify their contributions to the \textit{FairInt} framework.
In Table~\ref{tab:tb2}, the \textit{Vanilla FairInt} refers to the \textit{FairInt} framework without the two interaction-wise bias mitigation regularization IFC and FC, \textit{+ FC} refers to the \textit{Vanilla FairInt} with Fairness Constraint, and \textit{+ IFC} refers to the \textit{Vanilla FairInt} with Interaction Fairness Constraint.
Although the debiasing effects of the \textit{+ FC} are not as significant as the \textit{FairInt}, it can achieve the same level of $\Delta DP$ and $\Delta EO$ as the vanilla MLP models debiased by FC in all the four datasets.
Compared with the \textit{+ FC}, the \textit{+ IFC} more focuses on improving $\Delta EO$ than $\Delta DP$. 
The reason is that the implicit mitigation regularization IFC focuses on optimizing the latent representation rather than directly mitigating bias behaviors against the model predictions.
Therefore, when the \textit{FairInt} adopts the IFC with the FC, it can markedly improve the fairness with the lower $\Delta DP$ and $\Delta EO$ than only leverage one of the two bias regularization components.

\subsection{(Q4) Analysis of Sensitive Hyper-parameter}\label{sec:analyhp}
In this section, we study the impact of the hyper-parameter $\lambda_{IFC}$ and $\lambda_{FC}$ in the Eq.~\ref{eq:lossfunc} to answer the research question \textbf{Q4}.
We conduct the sensitivity analysis for both the two hyper-parameters on the Adult and Bank Marketing datasets.
To analyze the influence of $\lambda_{FC}$, we fix the best $\lambda_{IFC}$ to see the trend of AUC, $\Delta DP$, and $\Delta EO$ when changing $\lambda_{FC}$ on the two datasets, respectively.
As shown in Figure~\ref{fig:sensi_fc_ifc}, in the proposed \textit{FairInt}, $\lambda_{FC}$ is not sensitive to downstream task performances AUC.
As for the two fairness metrics $\Delta DP$ and $\Delta EO$, they will be improved when the $\lambda_{FC}$ increases, and the improvement will gradually converge to a certain level.
And to analyze $\lambda_{IFC}$, we fix the best $\lambda_{FC}$ to observe the trend of AUC, $\Delta DP$ and $\Delta EO$ when changing $\lambda_{IFC}$ on the Adult and Bank Marketing datasets, respectively.
According to the observations from Figure~\ref{fig:sensi_fc_ifc}, in the \textit{FairInt}, $\lambda_{IFC}$ is not sensitive to downstream task performance AUC.
At the same time, the best $\lambda_{IFC}$ can typically achieve the best $\Delta DP$ when reaching the best $\Delta EO$ on both Adult and Bank Marketing datasets.

\vspace{-0.2cm}
\subsection{(Q5) Key Observations on Interaction}\label{sec:observ}
One of the benefits of modeling feature interaction is that it provides better interpretability by observing the pattern of modeled feature interactions.
Therefore, in this section, we provide the key observations on the feature interactions, which refer to the attention weights $a_{s,k}$ calculated by Eq.~\ref{eq:attweight} in our proposed FairInt.
Here, we show the feature interactions between the sensitive and non-sensitive attributes on the Adult dataset, and we treat the \textit{FairInt w/o Both} as a biased model, the \textit{FairInt w/o FC} as a slightly fair model, the \textit{FairInt w/o IFC} as a fair model, and \textit{FairInt} as a fairer model.
The feature interactions of \textit{FairInt w/o Both}, \textit{FairInt w/o FC}, \textit{FairInt w/o IFC} and \textit{FairInt} are shown in the Figure~\ref{fig:att_wei}.
In the four figures, the yellow points represent the mean values of each attention weight between the sensitive attribute \textit{gender} and a non-sensitive attribute.
By comparing the feature interactions between biased and fair models, we conclude with the two factors of the feature interactions, which are variance and mean value.
Fair models have a lower variance of each feature interaction between sensitive and non-sensitive attributes, and a mean value of one feature interaction represents the correlation between the sensitive and the non-sensitive attribute.
For example, comparing the attention weights of FairInt, the fairest one among the four models, with FairInt (w/o Both), the most unfair one among the four models, the feature interactions between \textit{gender} and all other non-sensitive attributes have lower variances in the more fair model.
Also, the mean value of the feature interaction between \textit{gender} and \textit{relationship} is lower in the fairest model, which implies the fairer model treats \textit{relationship} as a less relevant attribute against \textit{gender}.





\vspace{-0.1cm}
\section{Conclusion and Future Works}
In this paper, we proposed FairInt, an assumption-free framework that automatically identifies and mitigates the biased feature interactions. Our framework doesn't need prior knowledge to identify the  related attributes in advance for mitigating the unfair model predictions. FairInt is composed of Sensitive Attribute Reconstructor, Bias Interaction Detection, and Interaction-wise Bias Mitigation, which aims to predict pseudo-sensitive attributes, model the information of identified bias feature interactions, and mitigate biased interaction with FC and IFC, respectively. Experiments on four real-world datasets demonstrate that FairInt can alleviate the unfair model predictions while maintaining the competitive classification performance. 
As for the future direction, we will explore the novel fairness constraint by limiting the variance of feature interaction, which implies the fairness extent of the proposed FairInt.


\bibliographystyle{acm}
\bibliography{paper}
\end{document}